\documentclass[runningheads]{llncs}

\usepackage[T1]{fontenc}

\usepackage{graphicx}

\usepackage{amsmath,amssymb}
\usepackage{color}
\usepackage{bm}
\usepackage{subcaption}
\usepackage{xcolor}

\begin{document}

\title{Learning Video-independent Eye Contact Segmentation from In-the-Wild Videos}
\titlerunning{Unconstrained Video-independent Eye Contact Segmentation}

\author{Tianyi Wu\orcidID{0000-0001-9077-5632} \and
Yusuke Sugano\orcidID{0000-0003-4206-710X}}

\authorrunning{T. Wu and Y. Sugano}

\institute{Institute of Industrial Science, The University of Tokyo\\
\email{\{twu223, sugano\}@iis.u-tokyo.ac.jp}
}

\maketitle              
\begin{abstract}
Human eye contact is a form of non-verbal communication and can have a great influence on social behavior.
Since the location and size of the eye contact targets vary across different videos, learning a generic video-independent eye contact detector is still a challenging task.
In this work, we address the task of one-way eye contact detection for videos in the wild.
Our goal is to build a unified model that can identify when a person is looking at his gaze targets in an arbitrary input video.
Considering that this requires time-series relative eye movement information, we propose to formulate the task as a temporal segmentation.
Due to the scarcity of labeled training data, we further propose a gaze target discovery method to generate pseudo-labels for unlabeled videos, which allows us to train a generic eye contact segmentation model in an unsupervised way using in-the-wild videos.
To evaluate our proposed approach, we manually annotated a test dataset consisting of $52$ videos of human conversations.
Experimental results show that our eye contact segmentation model outperforms the previous video-dependent eye contact detector and can achieve $71.88\%$ framewise accuracy on our annotated test set.
Our code and evaluation dataset are available at \url{https://github.com/ut-vision/Video-Independent-ECS}.

\end{abstract}

\section{Introduction}

\begin{figure}[t]
    \centering
    \includegraphics[width=1\linewidth]{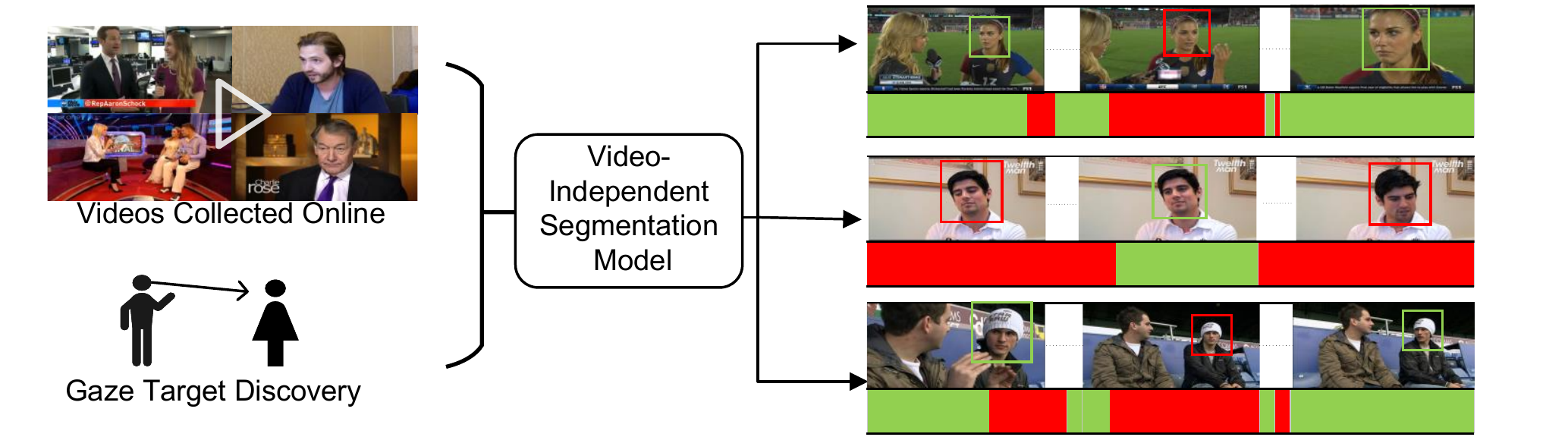}
    \caption{Illustration of our proposed task of video-independent eye contact segmentation. Given a video sequence and a target person, the goal is to segment the video into fragments of the target person having and not having eye contact with his potential gaze targets.}
    \label{fig:intro}
\end{figure}

Human gaze and eye contact have strong social meaning and are considered key to understanding human dyadic interactions.
Studies have shown that eye contact functions as a signaling mechanism~\cite{psy,eyetrackingsignaling}, indicates interest and attention~\cite{psy2,psy3}, and is related to certain psychiatric conditions~\cite{eyetrackingpersonality,asd,ADHD}.
The importance of human gazes in general has also been well recognized in the computer vision community, leading to a series of related research work on vision-based gaze estimation techniques~\cite{gaze_eye1,gaze_eye2,gaze_eye_yu2020,gaze_twoeyes1,gaze_twoeyes2,gaze_twoeyes3,appearancebasedgaze1,appearancebasedgaze2,appearancebasedgaze3,xgaze,appearancebased_reconstruction}.
Recent advances in vision-based gaze estimation have the potential to enable robust analyses of gaze behavior, including one-way eye contact.
However, gaze estimation is still challenging in images with extreme head poses and lighting conditions, and it is not a trivial task to robustly detect eye contacts in in-the-wild situations.

Several previous studies have attempted to directly address the task of detecting one-way eye contact~\cite{unsupervisedgaze17,unsupervisedgaze18,gazelocking,eyetracking_ye2015,Chong2017eyecontact}.
Given its binary classification nature, one-way eye contact detection can be a simpler task than regressing gaze directions.
However, unconstrained eye contact detection remains a challenge.
Fundamentally speaking, one-way eye contact detection is an ill-posed task if the gaze targets are not identified beforehand.
Fully supervised approaches~\cite{gazelocking,eyetracking_ye2015,Chong2017eyecontact} necessarily result in environment-dependent models that cannot be applied to eye contact targets with different positions and sizes.
Although there have been some work that address this task using unsupervised approaches that automatically detect the position of gaze targets relative to the camera~\cite{unsupervisedgaze17,unsupervisedgaze18}, they still require a sufficient amount of unlabeled training data from the target environment.
Learning a model that can detect one-way eye contact from arbitrary inputs independently of the environment is still a challenging task.

This work aims to address the task of unconstrained video-independent one-way eye contact detection. 
We aim to train a unified model that can be applied to arbitrary videos in the wild to obtain one-way eye contact moments of the target person without knowing his gaze targets beforehand.
Since the position and size of the eye contact targets vary from video to video, it is nearly impossible to approach this task frame by frame.
However, we humans can recognize when eye contact occurs from temporal eye movements, even when the target object is not visible in the scene.
Inspired by this observation, we instead formulate the problem as a segmentation task utilizing the target person's temporal face appearance information from the input video (Fig.~\ref{fig:intro}).
The remaining challenge here is that this approach requires a large amount of eye contact training data. 
It is undoubtedly difficult to manually annotate training videos covering a wide variety of environmental and lighting conditions.

To train the eye contact segmentation model, we propose an unsupervised gaze target discovery method to generate eye contact pseudo-labels from noisy appearance-based gaze estimation results.
Since online videos often contain camera movements and artificial edits, it is not a trivial task to locate eye contact targets relative to the camera. 
Instead of making a strong assumption about a stationary camera, we assume only that the relative positions of the eye contact target and the person are fixed. 
Our method analyzes human gazes in the body coordinate system and treats high-density gaze point regions as positive samples.
By applying our gaze target discovery method to the VoxCeleb2 dataset~\cite{voxceleb2}, we obtain a large-scale pseudo-labeled training dataset.
Based on the initial pseudo-labels, our segmentation model is trained iteratively using the original facial features as input.
We also manually annotated $52$ videos with eye contact segmentation labels for evaluation, and experiments show that our approach can achieve $71.88\%$ framewise accuracy on our test set and outperforms video-dependent baselines. 

Our contributions are threefold.
First, to the best of our knowledge, we are the first to formulate one-way eye contact detection as a segmentation task. 
This formulation allows us to naturally leverage the target person's face and gaze features temporally, leading to a video-independent eye contact detector that can be applied to arbitrary videos. 
Second, we propose a novel gaze target discovery method robust to videos in the wild. 
This leads to high-quality eye contact pseudo-labels that can be further used for both video-dependent eye contact detection and video-independent eye contact segmentation.
Finally, we create and release a manually annotated evaluation dataset for eye contact segmentation based on the VoxCeleb2 dataset.

\section{Related work}

\subsection{Gaze Estimation and Analysis}

\paragraph{Appearance-based Gaze Estimation}
Appearance-based gaze estimation directly regresses the input image into the gaze direction and only requires an off-the-shelf camera.
Although most of the work take the eye region as input~\cite{gaze_eye1,gaze_eye2,gaze_eye_yu2020,gaze_twoeyes1,gaze_twoeyes2,gaze_twoeyes3}, some demonstrated the advantage of using the full face as input~\cite{MPIIGAZE,appearancebasedgaze1,appearancebasedgaze2,appearancebasedgaze3,xgaze,appearancebased_reconstruction}. 
If the eye region is hardly visible, possibly due to low resolution, extreme head poses, and poor lighting conditions, the full-face gaze model can be expected to infer the direction of the human gaze from the rest of the face. 
Since most gaze estimation datasets are collected in controlled laboratory settings~\cite{xgaze,rtgene,eyediap}, in-the-wild appearance-based gaze estimation remains a challenge.
Some recent efforts have been made to address this issue by domain adaptation~\cite{collaborativeAdaption,simgan} or using synthetic data~\cite{appearancebased_reconstruction,gaze_synthesis_yu2019,sted}. 
Note that eye contact detection is a different task from gaze estimation and is still difficult even with a perfect gaze estimator due to the unknown gaze target locations.
The goal of this work is to improve the accuracy of eye contact detection on top of the state-of-the-art appearance-based gaze estimation method.

\paragraph{Gaze Following and Mutual Gaze Detection}
First proposed by Recasens~\textit{et~al}.~\cite{gazefollowing1}, gaze following aims to estimate the object where the person gazes in an image~\cite{gazefollowing2,gazefollowing_transformer,gazefollowing3,gazefollowing4,gazefollowing5,gazefollowing_outside1,gazefollowing_outside2}.
Another line of work is mutual gaze detection, which aims to locate moments when two people are looking at each other.
Mutual gaze is an even stronger signal than one-way eye contact in reflecting the relationship between two people~\cite{LAEO2,LAEO-NET,LAEO-NET++}. 
The problem of mutual gaze detection was first proposed by Marin-Jimenez~\textit{et~al}.~\cite{LAEO1}.
Our target differs from these tasks in two ways.
First, we are interested in finding the moments in which one-way eye contact occurs to gaze targets, rather than determining the location of gaze targets on a frame-by-frame basis or detecting mutual gazes.
Second, since our proposed method performs eye contact detection by segmenting the video based on the person's facial appearance, it can handle the cases where the gaze targets are not visible from the scene. 
Although some gaze following work~\cite{gazefollowing_outside1,gazefollowing_outside2} can tell when gaze targets are outside the image, most of them are designed with the implicit assumption that the gaze target is included in the image. 

\subsection{Eye Contact Detection}
Several previous works address the task of detecting eye contact specifically with the camera~\cite{gazelocking,eyetracking_ye2015,Chong2017eyecontact}. 
However, such pre-trained models cannot be applied to videos with the target person attending to gaze targets of different sizes and positions.
Recent progress in appearance-based gaze estimation allows unsupervised detection of one-way eye contacts in third-person videos using an off-the-shelf camera~\cite{unsupervisedgaze17,unsupervisedgaze18}.
Zhang~\textit{et~al}. assume a setting in which the camera is placed next to the gaze target and propose an unsupervised gaze target discovery method to locate the gaze target region relative to the camera~\cite{unsupervisedgaze17}.
They first run the appearance-based gaze estimator on all input sequences of human faces to get 3D gaze directions and then compute gaze points in the camera plane.
This is followed by density-based clustering, which identifies high-density gaze point regions as the locations of gaze targets.
Based on this idea, M\"{u}ller~\textit{et~al}.~studies eye contact detection in a group of $3$ - $4$ people having conversations~\cite{unsupervisedgaze18}.
Based on the assumption that all listeners would look at the speaker most of the time, they use audio clues to more accurately locate gaze targets in the camera plane. 

There are two major limitations that make these two approaches inapplicable to videos in the wild. 
First, in many online videos, camera movements and jump cuts are common, making the camera coordinate system inconsistent throughout the video. 
Meanwhile, since gaze points are essentially the intersection between the gaze ray and the plane $z = 0$ in the camera coordinate system, the gaze points corresponding to gaze targets far from the camera will naturally be more scattered than those corresponding to gaze targets close to the camera when receiving the same amount of eye gazes.
Consequently, density-based clustering would fail to identify potential gaze targets far from the camera on the camera plane.
Second, both works only explored video-dependent eye contact detection, \textit{i}.\textit{e}., training one model for each test video.
Instead, we study the feasibility of training a video-independent eye contact segmentation model that can be applied to different videos in the wild.

\subsection{Action Segmentation}
Action segmentation is the task of detecting and segmenting actions in a given video. 
Various research works have focused on designing the network architecture for the task.
Singh~\textit{et~al}.~\cite{actionsegmentation_bilstm} propose to feed spatial-temporal video representations learned by a two-stream network to a bi-directional LSTM to capture dependencies between temporal fragments. 
Lea~\textit{et~al}.~\cite{tcn} propose a network that performs temporal convolutions with the encoder-decoder architecture (ED-TCN). 
Recently, many works have tried to modify ED-TCN by introducing deformable convolutions~\cite{tdrn}, dilated residual layers~\cite{mstcn}, and dual dilated layers~\cite{mstcn++}.
In this work, we adopt MS-TCN++~\cite{mstcn++} as our segmentation model.

Since labeling action classes and defining their temporal boundaries to create annotations for action segmentation can be difficult and costly, some work explored unsupervised action segmentation~\cite{unsupervisedsegmentationGMM,unsupervisedsegmentationKNN,unsupervisedsegmentationnext,unsupervisedsegmentationshuffle,unsupervisedsegmentationautoencoder}.
Based on the observation that similar actions tend to appear in a similar temporal position in a video, most of these works rely on learning framewise representations through the self-supervised task of time stamp prediction~\cite{unsupervisedsegmentationKNN,unsupervisedsegmentationnext,unsupervisedsegmentationshuffle,unsupervisedsegmentationautoencoder}.
However, it is difficult to apply these methods directly to our scenario because eye contact is a sporadic activity that can occur randomly over time.
We instead leverage human gaze information and deduce the gaze target position from gaze point statistics.

\section{Proposed Method}

\begin{figure}[t]
    \centering
    \includegraphics[width=1.\linewidth]{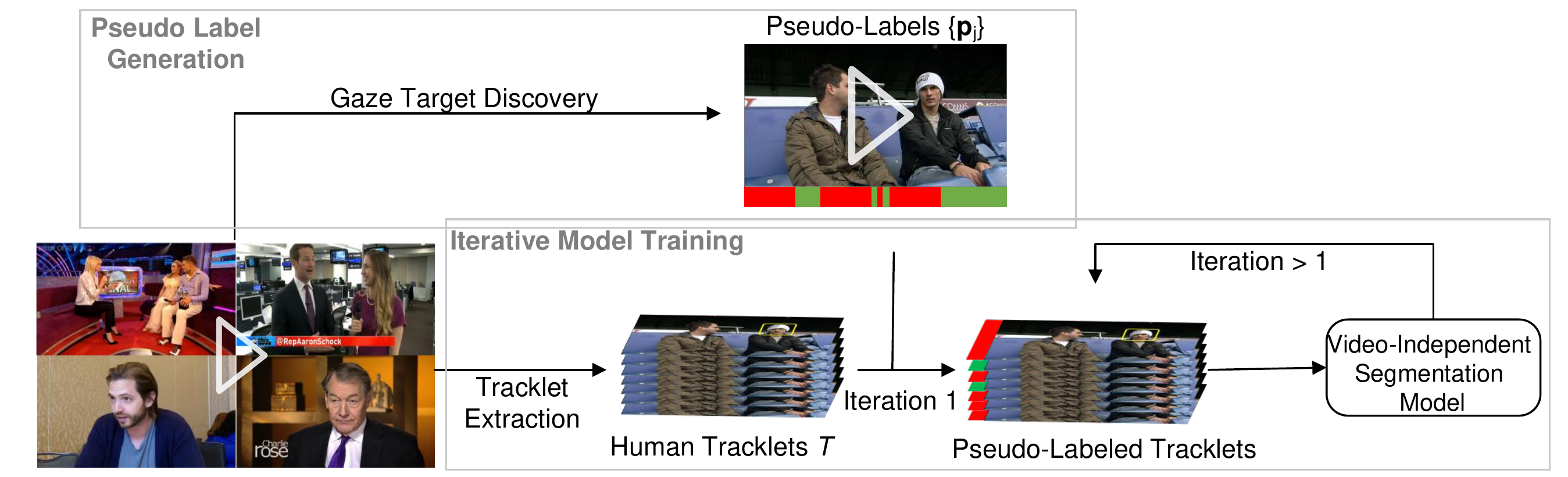}
    \caption{An overview of our proposed unsupervised training pipeline for video-independent eye contact segmentation.} 
    \label{fig:overview}
\end{figure}

Our proposed eye contact segmentation network takes input a \emph{tracklet}, \textit{i}.\textit{e}., a sequence of video frames in which the target person is tracked and outputs framewise eye contact predictions.
Formally, given a tracklet with $I$ frames $T = \{\bm{I}_i\}_{i=1}^I$, our objective is to train a model to produce framewise binary predictions of one-way eye contacts $Y = \{\bm{y}_i\}_{i=1}^I$ of the person, where $\bm{y}_i \in [0,1]^2$ is a two-dimensional one-hot vector.
We define gaze targets as physical targets with which the person interacts, such as the camera and another person in the conversation.
These gaze targets do not have to be visible in the video.

Fig.~\ref{fig:overview} shows an overview of the proposed unsupervised approach to train the segmentation network. 
Our method consists of two stages, \textit{i}.\textit{e}.,  \emph{pseudo-label generation} and \emph{iterative model training}. 
We start by collecting a large set of $M$ unlabeled conversation videos $\mathcal{V} = \{V_m\}_{m=1}^M$ from online.
For each video $V_m$ with $J_m$ frames, we first generate framewise pseudo-labels $ \{\bm{p}_{j}\}_{j=1}^{J_m}$, where $\bm{p}_{j} \in [0,1]^2$, using our proposed method of gaze target discovery. 
We also track the target person to extract a set of tracklets $\{T_{k}\}_{k=1}^{K_m}$ from each video $V_m$.
The pseudo-labels $ \{\bm{p}_{j}\}$ are also split and assigned to each corresponding tracklet as a tracklet-wise set of pseudo-labels $P_k$.
The collection of all tracklets $\mathcal{T} = \{T_{n}\}_{n=1}^{N}$ obtained from $\mathcal{V}$, where $N=\sum_m K_m$, and their corresponding collection of pseudo-labels $\mathcal{P} = \{P_n\}_{n=1}^N$ are then used to train our eye contact segmentation model. 
Since our proposed gaze target discovery does not leverage temporal information, we propose an iterative training strategy that iteratively updates the pseudo-labels using the trained segmentation model that has learned rich temporal information.
In the following sections, we describe details of our pseudo-label generation and iterative model training processes.

\subsection{Pseudo Label Generation}
\label{pl_generator}

\begin{figure}[t]
    \centering
    \includegraphics[width=1.\linewidth]{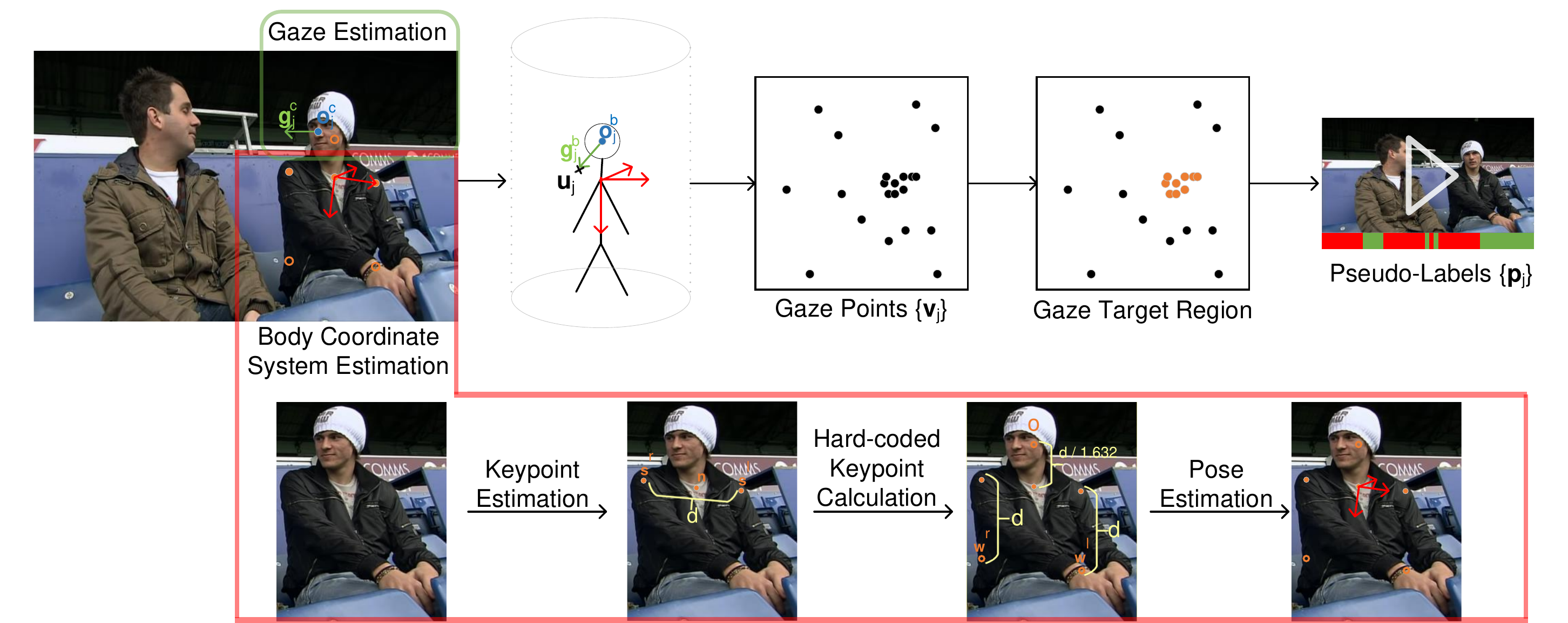}
    \caption{An overview of the pseudo-label generation stage using our proposed method of gaze target discovery.} 
    \label{fig:cylinder}
\end{figure}

We generate framewise pseudo-labels $\{\bm{p}_{j}\}$ for each training video $V_m$ using our proposed gaze target discovery, which automatically locates the position of the gaze targets in the body coordinate system.
An overview is illustrated in Fig.~\ref{fig:cylinder}.
In a nutshell, our proposed gaze target discovery obtains the target person's 3D gaze direction in the body coordinate system and identifies the high-density gaze regions.
Since eye contact targets tend to form dense gaze clusters, these gaze regions are treated as potential gaze target locations.
In the following sections, we give details of our proposed gaze target discovery and body pose estimation.

\subsubsection{Gaze Target Discovery} \label{fixations}
For each frame $\bm{I}_j$ of the video $V_m$, we run an appearance-based gaze estimator to obtain the gaze vector $\bm{g}^{c}_{j}$ of the person of our interest in the camera coordinate system.
Through the data normalization process for gaze estimation~\cite{appearancebasedgaze1}, we also obtain the center of the face $\bm{o}^{c}_{j}$ as the origin of the gaze vector.
We also perform body pose estimation to obtain the translation vector $\bm{t}_{j}$ and the rotation matrix $\bm{R}_{j}$ from the body coordinate system to the camera coordinate system.
$\bm{g}^{c}_{j}$ and $\bm{o}^{c}_{j}$ are then transformed from the camera coordinate system to the body coordinate system through $\bm{g}^{b}_{j} =  \bm{R}_j^{-1} \bm{g}^{c}_{j}$ and $\bm{o}^{b}_{j} =  \bm{R}_{j}^{-1} (\bm{o}^{c}_{j} - \bm{t}_j)$,
where $\bm{o}^{b}_{j}$ indicates the face center in the body coordinate system.
Therefore, $\bm{o}^{b}_{j} + \beta \bm{g}^{b}_{j}$ defines the gaze ray in the body coordinate system.

For each video $V_m$, we then compute a set of intersection points $\{\bm{u}_{j}\}$ between the gaze rays and a cylinder centered at the origin of the body coordinate system with a radius of $r$ and convert these 3D intersection points to 2D gaze points $\{\bm{v}_{j}\}$ on the cylinder plane by cutting the cylinder along the line $(0, y, r)$ parameterized with $y$.
%, that is located at the z-axis of the body coordinate system points opposite to the chest-facing direction.
We apply OPTICS clustering~\cite{optics} on the set of 2D gaze points $\{\bm{v}_{j}\}$ and treat the identified clusters as eye contact regions for the $m$-th video $V_m$. 
We generate pseudo-labels $\{\bm{p}_{j}\}$ for $V_m$ by treating all identified eye contact regions as positive samples and others as negative samples.

\subsubsection{Body Pose Estimation} 
\label{pose}
We estimate the 3D body pose $\bm{R}$ and $\bm{t}$ of the target person based on the 3D model fitting.
We define our six-point 3D generic body model according to the average human body statistics~\cite{humanbody}, consisting of nose, neck, left and right shoulder, and left and right waist keypoints.
This 3D body model is in a right-handed coordinate system, which means that the chest-facing direction is the negative Z-axis direction. 
Given the corresponding 2D keypoint locations from the target image, we can fit the 3D model using the P6P algorithm~\cite{epnp} assuming (virtual) camera parameters.
This gives us the translation vector $\bm{t}$ and the rotation matrix $\bm{R}$ from the body coordinate system to the camera coordinate system. 

To locate the six 2D body keypoints, we rely on a pre-trained 2d keypoint-based pose estimator.
For each frame, the pose estimator is expected to take the whole frame as input, and output body keypoints including the ones corresponding to our six-point body model.
However, directly using the six 2D keypoints from the pose estimator can lead to inconsistent results throughout frames.
This inconsistency arises from the fact that there normally exist at least two near-optimal solutions with similar reprojection errors due to the symmetric nature of the human body.
Therefore, we introduce some subtle asymmetry in the 3D body model and stabilize the pose by introducing hard-coded keypoints, as illustrated in the lower part of Fig.~\ref{fig:cylinder}.
Specifically, we only use three keypoints that correspond to the left shoulder $\bm{s}^l$, the right shoulder $\bm{s}^r$, and the neck $\bm{n}$ from the pose estimator.
We calculate the other three keypoints, the left waist $\bm{w}^l$, the right waist $\bm{w}^r$, and the nose $\bm{e}$, assuming that the target person is standing straight.
The keypoints of the waist are defined as $\bm{w}^l = \bm{s}^l + ( 0, d )^{T}$ and $\bm{w}^r = \bm{s}^r + ( 0, d )^{T}$, where $d = |\bm{s}^r - \bm{s}^l|_2$ indicates the length of the shoulder, and the nose is defined as $\bm{e} = \bm{n} - ( 0, \alpha d )^{T}$.
We also set the ratio $\alpha = 1.632$ according to the same statistics of the human body~\cite{humanbody}.

\subsection{Iterative Model Training}

\begin{figure}[t]
    \centering
    \includegraphics[width=1.\linewidth]{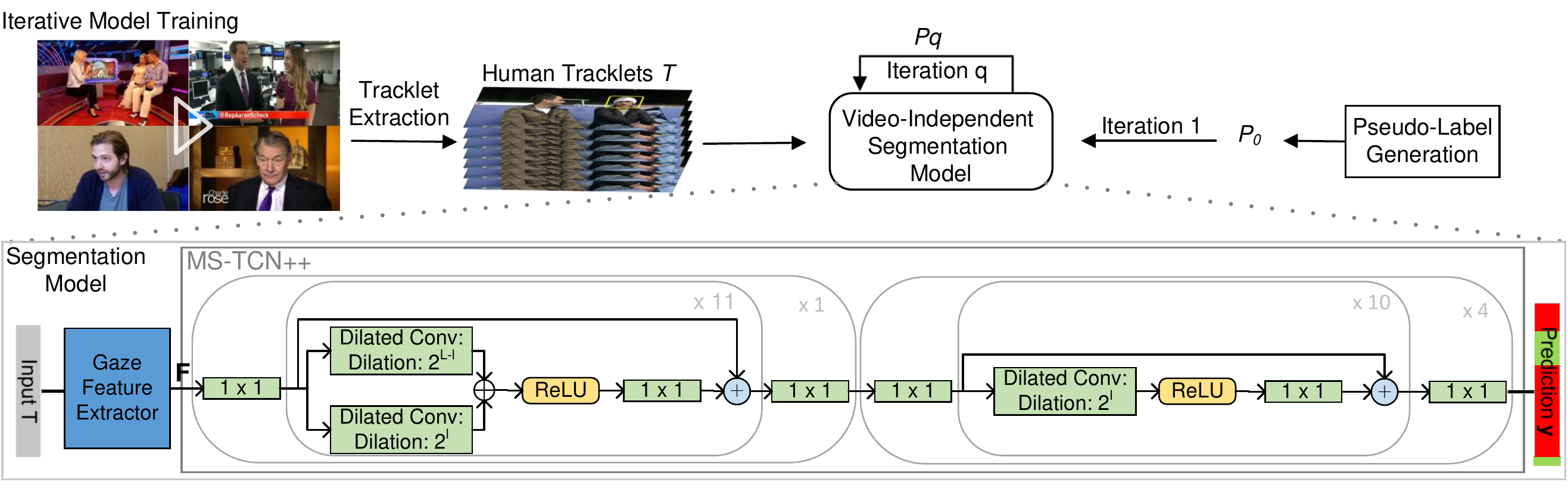}
    \caption{An overview of the iterative model training stage with an illustration for our network structure.}
    \label{fig:MSTCN++}
\end{figure}

By applying gaze target discovery to all videos in $\mathcal{V}$ and extracting tracklets, we obtain an initial training dataset consisting of $\mathcal{T}$ and $\mathcal{P}$.
However, our proposed gaze target discovery process generates noisy pseudo-labels.
Consequently, the initial labels $\mathcal{P}$ obtained from appearance-based gaze estimation results can be oversegmented, violating the nature of eye contact segmentation.
To address this problem, we propose to iteratively train new segmentation models supervised by the pseudo-labels $\mathcal{P}_q$ generated from the model trained in the previous iteration.
Our segmentation model is trained to take the low-level gaze CNN feature as input and is expected to auto-correct the initial noisy pseudo-labels by attending to temporal information through the iterative process.

Fig.~\ref{fig:MSTCN++} shows an overview of the iterative model training stage.
Our segmentation network is based on the MS-TCN++ architecture~\cite{mstcn++} that takes as input a tracklet $T_n$ and outputs its frame-wise eye contact predictions $\bm{y}_n$.
At iteration 1, the model is supervised with $\mathcal{P}_1 = \mathcal{P}$ generated from the pseudo-label generator. 
In every subsequent iteration $q > 1$, the model will be supervised with better pseudo-labels $\mathcal{P}_1$ than the models trained before and could learn richer temporal information.
We repeat this process to max $Q$ iterations.

\subsubsection{Network Architecture}

The structure of the segmentation network is illustrated in the lower part of Fig.~\ref{fig:MSTCN++}.
For each frame in the input tracklet $T$ of length $I$, we extract and normalize the face image of the target person according to~\cite{appearancebasedgaze1}.
It is then fed to a pre-trained gaze estimation network based on the ResNet architecture~\cite{resnet} with 50 layers followed by a fully connected regression head.
We extract the gaze feature vectors $\bm{f}_i \in \mathbb{R}^{2048}$ from the last layer and concatenate all gaze features $\{\bm{f}_i\}$ collected from the tracklet along the temporal dimension to form the gaze feature matrix $\bm{F} \in \mathbb{R}^{2048 \times I}$, which will be used as input to the segmentation block.

Based on the MS-TCN++ architecture~\cite{mstcn++}, the segmentation block consists of a prediction stage and several refinement stages stacked upon the prediction stage. 
The prediction stage has $11$ dual dilated layers.
At the $l$-th dual dilated layer, the network performs two dilated convolutions with dilation rates $l$ and $11-l$.
The features after the two dilated convolutions are first concatenated, so that the network is able to attend to long-range temporal information even at the early stage.
This is then followed by ReLU activation, pointwise convolution and skip connections.
The refinement stages are similar to the prediction stage, except that the 11 dual dilated layers are replaced with 10 dilated residual layers. 

For the loss function, we follow the original paper and use a combination of cross-entropy loss $\mathcal{L}_{\textrm{cls}}$ and truncated mean squared loss $\mathcal{L}_{\textrm{t-mse}}$~\cite{mstcn++}.
$\mathcal{L}_{\textrm{t-mse}}$ is defined based on the difference in the prediction between adjacent frames and encourages smoother model prediction. 
The loss for a single stage is defined as $\mathcal{L}_s = \frac{1}{|B|}\sum_{b \in B}(\mathcal{L}_{\textrm{cls}}(\bm{p}_b, \bm{y}_b) + \lambda \mathcal{L}_{\textrm{t-mse}}(\bm{y}_b))$, 
where $B$ indicates the training batch, and $\lambda$ is a hyperparameter controlling the extent of $\mathcal{L}_{\textrm{t-mse}}$. 
Finally, the overall loss for all stages is the sum of $\mathcal{L}_s$ at each stage.

\subsubsection{Tracklet Extraction} \label{tracklets}
To extract tracklets, we run face detection and face recognition on each video in $\mathcal{V}$ in the training dataset.
A tracklet $T_k$ is formed only if the IoU between the bounding boxes of the faces of consecutive frames is greater than $\tau_{\mathrm{IoU}}$. 
The tracklet is also disconnected if the neck and shoulder keypoints cannot be detected.
We also discard short tracklets that do not exceed 4 seconds.

Since pseudo-labels extracted from the gaze target discovery are person-specific, we also need to make sure that the training tracklets are extracted from the specific target person.
To this end, we add the cosine similarity threshold $\tau_{c}$ to construct training tracklets.
Assuming that a set of reference face images is given, we compute the cosine similarity of the detected face with each of the reference faces.
As long as one of them is greater than $\tau_{c}$, the tracklet continues.
Note that this threshold is not required during inference.

\subsection{Implementation Details}

The gaze estimation network is pre-trained on the ETH-XGaze dataset~\cite{xgaze}, and we follow their work to perform face detection, face normalization, and gaze estimation.
We use OpenPose~\cite{openpose} as our 2d keypoint pose estimator. 
For face recognition, we use ArcFace~\cite{arcface}, and set the cosine similarity threshold $
\tau_{c} = 0.4$ and the IoU threshold $\tau_{\mathrm{IoU}} = 0.4$. 

During the pseudo-label generation stage, we set the radius of the cylinder $r = 1000mm$. 
We also noticed that the hyperparameter of OPTICS can significantly affect the pseudo-label quality. 
In particular, we found that smaller max epsilon values should be given to longer videos, as the clustering space of long videos is much denser than that of short videos. 
To address this issue, we first set the max epsilon to $8$ and perform OPTICS clustering. 
If no clusters are found, we continue to increment the max epsilon until at least one cluster is found.

During training, we split the pseudo-labeled dataset into training and validation splits with a ratio of $8:2$.
To train MS-TCN++, we follow the suggested hyperparameters for the network architecture. 
We use Adam~\cite{adams} optimizer and set the learning rate to $0.0005$. 
We did not use dropout layers, and for the loss function, we set $\lambda=0.15$.
We empirically set the maximum number of iterations $Q = 4$ and trained 50 epochs for each iteration.

\section{Experiments}

\subsection{Dataset} \label{tab:dataset}
We build our dataset upon the VoxCeleb2 dataset~\cite{voxceleb2}, which consists of celebrities being interviewed or speaking to the public.
The trimmed VoxCeleb2 dataset includes short video clips only containing the head crops of celebrities and has been used in various tasks including talking head generation~\cite{talking_head_generation} and speaker identification~\cite{voxceleb2}.
Since our method also requires body keypoints detection, we opt to process the raw videos used in the VoxCeleb2 dataset.

During training, we use randomly selected videos from celebrities id00501 to id09272. 
Due to the online availability and the computational cost of tracklet extraction, we used 5\% of the entire dataset for training.
After downloading the raw videos, we converted them to $25$ fps. 
If the video is too short, there will not be enough gaze points to reliably identify high-density regions.
On the other hand, if the video is too long, the possibility of the video containing multiple conversation sessions becomes high.
Therefore, we only used videos of a duration between $2$ and $12$ minutes for training. 
In total, we pre-processed $4926$ raw videos.
During tracklet extraction, we obtain the sets of reference images from the VGGFace2 dataset~\cite{vggface2}. 
For each celebrity, we only used the first $30$ face images.
This results in $49826$ tracklets, which is equivalent to roughly $177.7$ hours. 

\begin{figure}[t]
    \centering
    \includegraphics[width=1\linewidth]{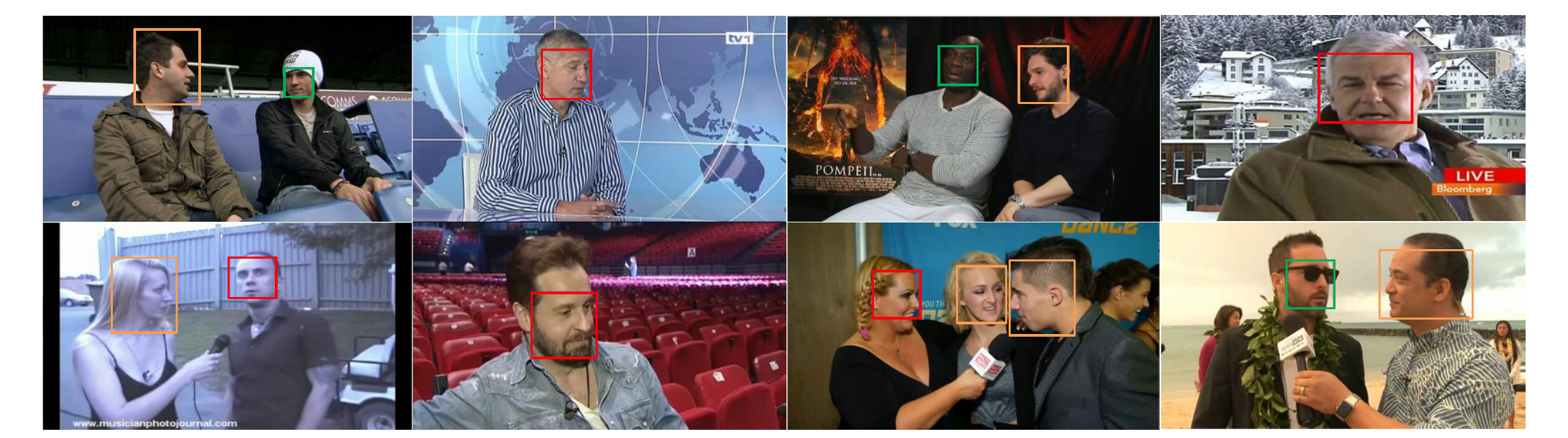}
    \caption{Some example video frames randomly selected from our test dataset. We show the target celebrities in green and red bounding boxes, with green and red indicating positive and negative ground-truth labels. If visible, their gaze targets are also shown in orange bounding boxes.}
    \label{fig:dataset}
\end{figure}

To evaluate our method, we manually annotate $52$ videos (summing up to $3.6$ hours) from celebrity id00016 to id00500 using ELAN~\cite{elan}. 
Each video is selected from a different celebrity to ensure identity diversity. 
We treat the host, camera, and other interviewees who interacted with the target celebrity as eye contact targets, meaning that there can be multiple eye contact targets in each video. 
Note that these gaze targets do not necessarily need to be visible in the scene. 
In addition, some test videos have poor lighting conditions, extreme head poses, and low resolution, making them difficult to segment.
Fig.~\ref{fig:dataset} shows some randomly sampled frames from the test videos with the celebrity and gaze targets that we identified highlighted in the boxes.
In total, $48.14\%$ of the frames are labeled positive. 
We further examine the quality of the labels by visualizing gaze positions on the cylinder plane.
If the gaze target looks too scattered on the cylinder plane, we re-annotate the video.   
After forming tracklets from the test videos, we get $510$ tracklets ($1.9$ hours in total) with $48.12\%$ of the frames labeled positive.
These test tracklets are used as our test set to evaluate our proposed eye contact segmentation model.

\subsection{Evaluation}

We compare our method with video-dependent baselines and show the effectiveness of our design choices through ablation studies. 
Following previous work on action segmentation, we evaluate our proposed method using framewise accuracy, segmental edit score, and F1 scores at the overlap threshold $10\%$, $25\%$, and $50\%$, denoted by $F1@\{10,25,50\}$. 
Framewise accuracy measures the performance in the most straightforward way, but it favors longer tracklets and does not punish oversegmentation. 
The segmented F1 score reflects the degree of oversegmentation, and the F1 scores measure the quality of the prediction. 

\begin{table*}[t]
\caption{\label{tab:result} Comparison between our video-independent eye contact segmentation model and video-dependent baseline approaches.}
\begin{tabular*}{\textwidth}{c @{\extracolsep{\fill}} cccccc}
  \hline
  Method & Accuracy & Edit & F1@0.1 & F1@0.25 & F1@0.5\\ 
  \hline\hline
  CameraPlane + SVM~\cite{unsupervisedgaze17} & 57.72\% & 46.83 & 38.37 & 30.08 & 20.48\\ 
  \hline
  CylinderPlane + SVM & 68.52\% & 46.84 & 43.3 & 38.16 & 29.49\\ 
  \hline
  \hline
  Ours & \textbf{71.88\%} & \textbf{57.27} & \textbf{61.59} & \textbf{54.67} & \textbf{41.03}\\ 
  \hline
\end{tabular*}
\end{table*}

\subsubsection{Performance Comparison}

Table.~\ref{tab:result} shows the comparison of our proposed unsupervised video-independent eye contact segmentation model with unsupervised video-dependent eye contact detection baselines.  
The first row (\emph{CameraPlane + SVM}) is the re-implementation of the unsupervised method of Zhang~\textit{et~al}.~\cite{unsupervisedgaze17} that extracts framewise pseudo-labels in the camera plane and trains an SVM-based eye contact classifier based on a single frame input.
We did not set a safe margin around the positive cluster to filter out unconfident negative gaze points because we observed that it does not work well in in-the-wild videos, especially when there exist multiple gaze targets.
The second row (\emph{CylinderPlane + SVM}) applies our proposed cylinder plane gaze target discovery method, and an SVM is trained on the resulting pseudo-labels. 
Note that these two methods are video-dependent approaches, \textit{i}.\textit{e}., the models are trained specifically on the target video.
Our proposed cylinder plane achieves $68.52\%$ accuracy in video-dependent eye contact detection, outperforming the camera plane baseline by $10.8\%$, indicating the advantage of our pseudo-label generator in in-the-wild videos. 
The last row (\emph{Ours}) corresponds to the proposed unsupervised eye contact segmentation approach with an iterative training strategy.
Our method achieves $71.88\%$ framewise accuracy, outperforming the video-dependent counterpart (\emph{CylinderPlane + SVM}) by $3.36\%$ and the camera-plane baseline by $14.16\%$. 
It also achieves the highest segmented edit scores and F1 scores, indicating better segmentation qualities.

\subsubsection{Ablation Studies}

\begin{table*}[t]
\caption{\label{tab:ablation} Ablation results on our design choices.}
\begin{tabular*}{\textwidth}{c @{\extracolsep{\fill}} cccccc}
  \hline
  Method & Accuracy & Edit & F1@0.1 & F1@0.25 & F1@0.5\\ 
  \hline\hline
  CameraPlane~+~Generic SVM & 61.28\% & 44.34 & 39.32 & 32.44 & 22.46\\ 
  \hline  
  CylinderPlane~+~Generic SVM & 58.83\% & 39.86 & 34.26 & 27.19 & 18.08\\ 
  \hline
  CameraPlane~+~Ours & 65.55\% & 46.27 & 44.88 & 38.56 & 27.85\\ 
  \hline\hline
  Ours (1 iteration) & 70.04\% & 42.21 & 43.97 & 38.55 & 27.42\\ 
  \hline
  Ours (2 iteration) & 71.15\% & 50.60 & 55.09 & 48.80 & 36.54\\ 
  \hline
  Ours (3 iteration) & 71.41\% & 52.24 & 57.26 & 50.72 & 37.45\\ 
  \hline
  Ours (4 iteration) & \textbf{71.88\%} & \textbf{57.27} & \textbf{61.59} & \textbf{54.67} & \textbf{41.03}\\ 
  \hline
\end{tabular*}
\end{table*}

We also conduct ablation studies to show the effectiveness of our design choices, \textit{i}.\textit{e}, our problem formulation, our proposed gaze target discovery method and iterative training. 
The result is shown in Table.~\ref{tab:ablation}. 
\emph{CameraPlane + Generic SVM} is the baseline method that obtains pseudo-labels for tracklets using the gaze target discovery method of Zhang~\textit{et~al}.~\cite{unsupervisedgaze17} and trains an SVM-based generic video-independent eye contact detector. 
We choose to use SVM as the classifier simply for comparison with video-dependent baseline approaches, and SVM is trained through online learning optimized by SGD.
\emph{CylinderPlane + Generic SVM} replaces the gaze target discovery method of Zhang~\textit{et~al}.~\cite{unsupervisedgaze17} with our proposed gaze target discovery method and, therefore, is also a video-independent eye contact detection approach.
Although the baseline camera-plane approach outperforms our proposed cylinder-plane approach by $2.45\%$, both SVM-based detection models achieve framewise accuracy only slightly better than chance.
\emph{CameraPlane + Ours} obtains pseudo-labels on the camera plane but replaces the SVM with our segmentation model, making it a video-independent eye contact segmentation method.
It outperforms its detection counterpart by $4.27\%$, showing the superiority of our problem formulation.

Our proposed method without iterative training (\emph{Ours (1 iteration)}) achieves $70.04\%$ accuracy, showing the effectiveness of our proposed gaze target discovery when applied in the segmentation task setting.
The last three rows of Table.~\ref{tab:ablation} shows the effectiveness of the proposed iterative training strategy. 
From iteration $1$ to iteration $2$, we observed a decent improvement in model performance in terms of both accuracy and segmentation quality.
There are also gradual model improvements even in subsequent iterations, indicating its effectiveness.
Our proposed method in the last iteration outperforms that of the first iteration by $1.84\%$ in accuracy, and there is also a great improvement in the edit score and the F1 scores.

\subsubsection{Detailed Analysis and Discussions}

\begin{table}[t]
  \centering
   \caption{\label{tab:trackletlength} Evaluation of the model performance with different input length.}
\begin{tabular*}{\textwidth}{c @{\extracolsep{\fill}} cccccc}
\hline
 Length [s] & Accuracy & Edit & F1@0.1 & F1@0.25 & F1@0.5   \\ \hline \hline
\multicolumn{1}{c}{$> 4$} & 71.88\% & 57.27 & 61.59 & 54.67 & 41.03 \\ \hline
\multicolumn{1}{c}{$> 10$} & 74.27\% & 62.31 & 63.93 & 57.45 & 43.76 \\ \hline
\multicolumn{1}{c}{$> 30$} & 77.31\% & 63.76 & 64.86 & 58.60 & 45.96 \\  \hline
\multicolumn{1}{c}{$> 60$} & 85.14\% & 68.96 & 76.65 & 73.57 & 65.01 \\  \hline
\end{tabular*}%
\end{table}%

During both training and testing time, we use tracklets for more than 4 seconds as input to the model.
In Table~\ref{tab:trackletlength}, we vary the length of the input tracklet during the test time and analyze its effect on the performance of the model. 
As can be seen, the model performance improves as we increase the input tracklet length.
In particular, if the input length is more than one minute, our proposed method reaches an accuracy of $85.14\%$.

\begin{figure}[t]
     \centering
     \begin{subfigure}[t]{0.4\textwidth}
         \centering
         \includegraphics[width=\textwidth]{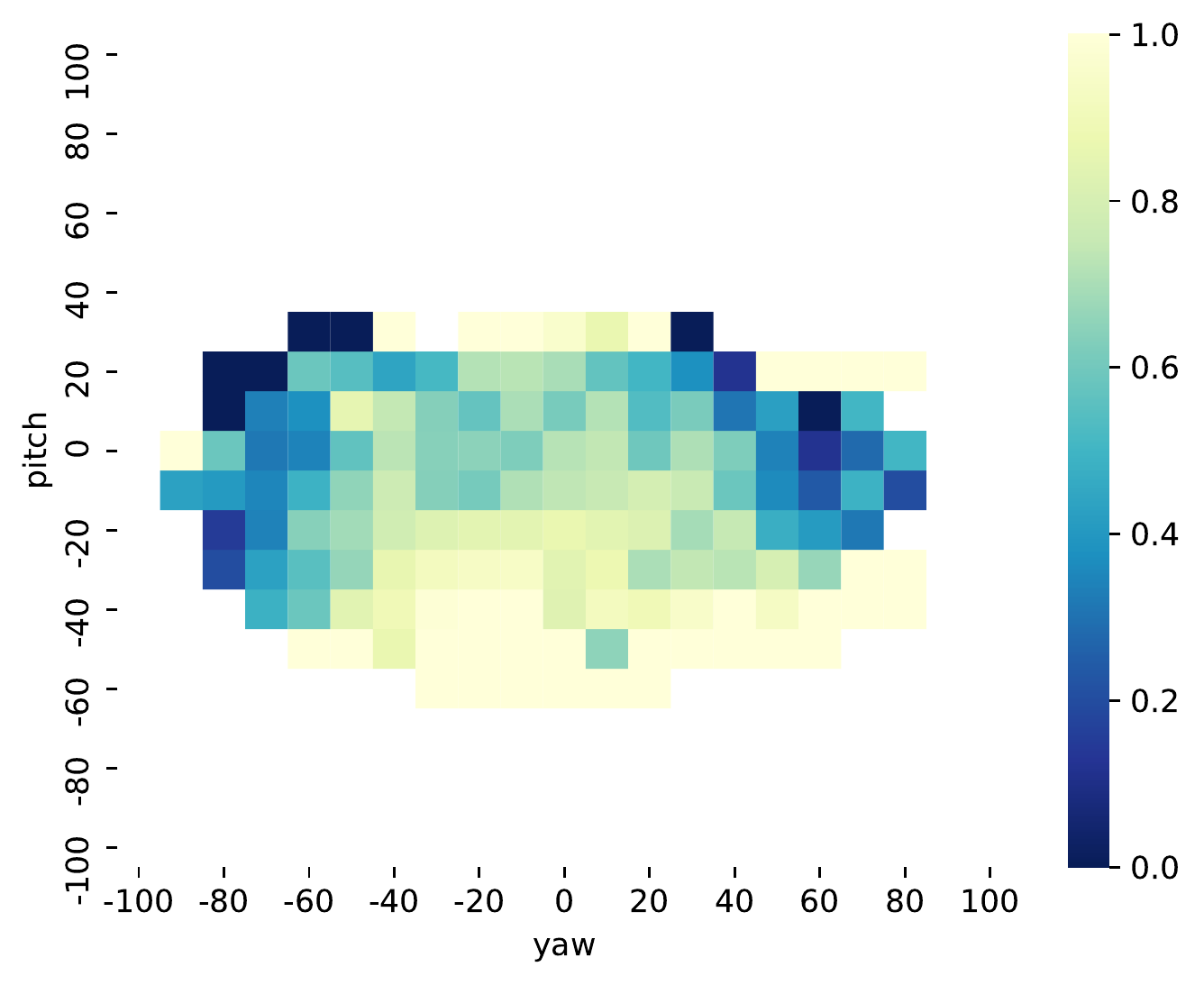}
         \caption{Model performance}
         \label{fig:headpose}
     \end{subfigure}
     \begin{subfigure}[t]{0.5\textwidth}
         \centering
         \includegraphics[width=\textwidth]{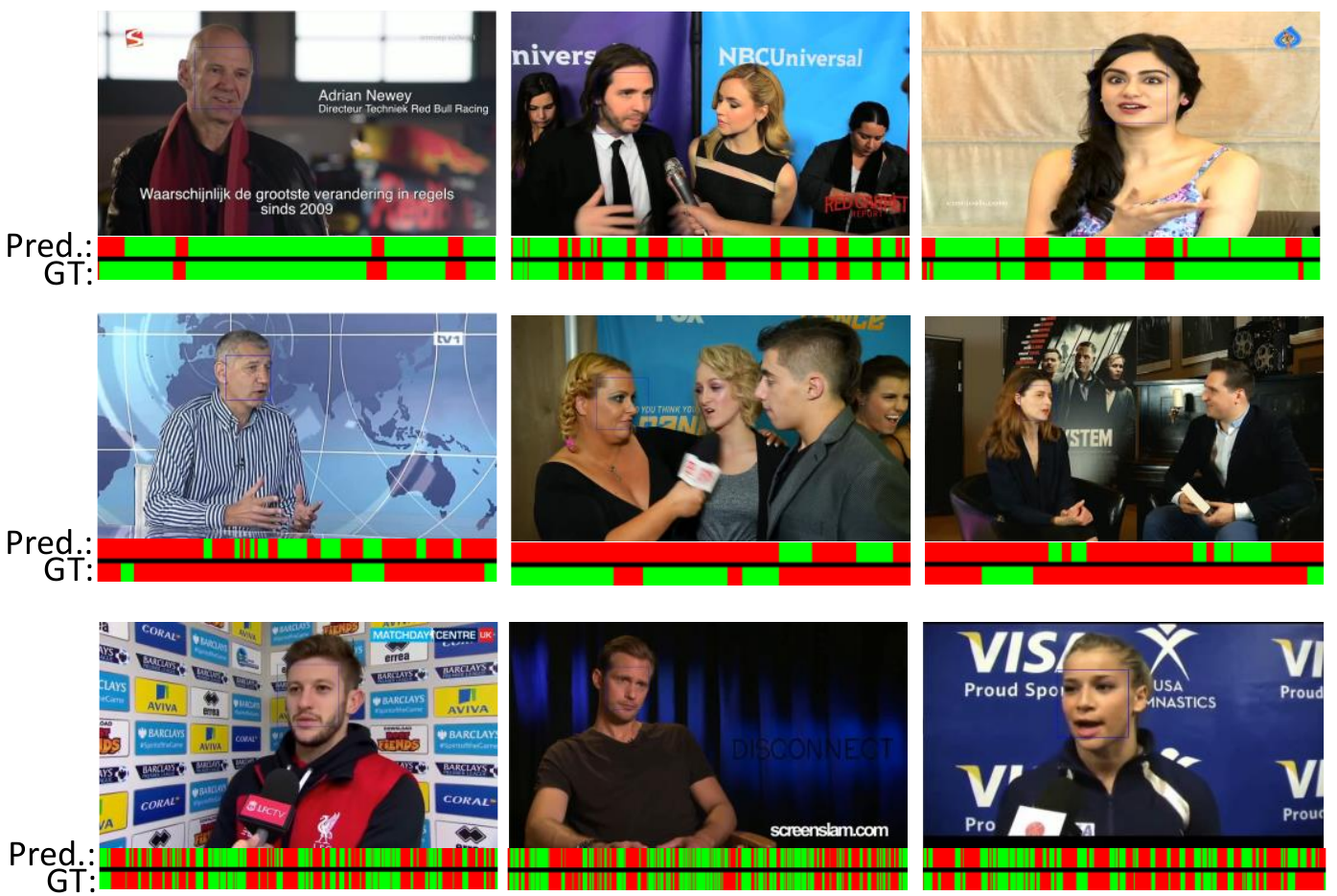}
         \caption{Qualitative results}
         \label{fig:examples}
     \end{subfigure}
     \caption{Accuracy of our segmentation model on different head pose ranges. We (a) visualize segmentation accuracy for each yaw-pitch interval bin, and (b) show examples of the model performance on test tracklets with a thumbnail image followed by model predictions and groudtruth annotations.}
\end{figure}

The higher performance observed in longer tracklets can also be attributed to the performance of the model in different head poses.
During the tracklet formation stage, we use face recognition to filter celebrity faces.
Since face recognition performance is limited on profile faces, tracklets containing extreme head poses tend to be much shorter than those containing only frontal faces.
During training, the lack of long tracklets with profile faces prevents the model from modeling long-term eye contact dependencies, leading to degraded performance on tracklets that contain mainly profile faces. 
The longer tracklets in the test set also contain mainly frontal faces, resulting in higher accuracy.

In Fig.~\ref{fig:headpose}, we visualize the accuracy of our trained model conditioned on different head poses.
We use HopeNet~\cite{Ruiz18headpose} to extract head poses of celebrity faces in each frame of the test tracklets and compute framewise accuracy for all yaw-pitch intervals.
We can observe that our model works the best when the pitch is between $-40$ and $-60$ degrees, \textit{i}.\textit{e}., looking downward. 
It can also achieve decent performance when both yaw and pitch are around $0$ degrees. 
However, when the yaw is lower than $-60$ degrees or higher than $60$ degrees, our model is even worse than random chance.

In Fig.~\ref{fig:examples}, we show some qualitative visualizations. 
We randomly present test tracklets with frontal faces in the first row and test tracklets with profile faces in the second. 
Although our model has decent performance on frontal-face tracklets, the predictions on profile-face tracklets seem almost random. 
We also present test tracklets longer than one minute in the third row, and all of these long tracks contain frontal faces.
Consequently, we can observe mainly fine-grained eye contact predictions in these examples.

Another limitation of our proposed approach lies in our full-face appearance-based estimator.
Fundamentally, the line of work on full-face appearance-based gaze estimation regresses the face images into gaze directions in the normalized camera coordinate system. 
Consequently, the gaze features used as input to the segmentation model only have semantic meaning in the normalized space, but not in the real camera space. 
In addition, the full-face appearance-based estimator trained on the gaze dataset collected in controlled settings tends to have limited performance in unconstrained images, especially when the person has extreme head poses unseen in the training dataset. 
This may also be a reason for the low performance of the model on the face of the profile.

Finally, our approach cannot handle the cases of moving gaze targets and humans.
Our gaze target discovery assumed fixed relative positions between the eye contact target and the person and would consequently give incorrect pseudo-labels in such cases.
Eye contact detection with moving gaze targets is a more challenging task than that with stable gaze targets.
We argue that in this case gaze information will not be sufficient for the model.
Information about the spatial relationship between the person and gaze targets should be introduced.

\section{Conclusion}
In this paper, we proposed and challenged the task of video-independent one-way eye contact segmentation for videos in the wild. 
We proposed a novel method of gaze target discovery to obtain frame-wise eye contact labels in unconstrained videos, which allows us to train the segmentation model in an unsupervised way.
By manually annotating a test dataset consisting of $52$ videos for evaluation, we showed that our proposed method can lead to a video-independent eye contact detector that can outperform previous video-dependent approaches and is especially robust for non-profile face tracklets.

\section*{Acknowledgement} This work was supported by JST CREST Grant Number JPMJCR1781.

\newpage
\bibliographystyle{splncs04}
\bibliography{mybibliography}

\end{document}